\documentclass[11pt]{article}

\usepackage[final]{acl}

\usepackage{times}
\usepackage{latexsym}

\usepackage[T1]{fontenc}

\usepackage[utf8]{inputenc}

\usepackage{microtype}
\usepackage{amsmath}
\usepackage{enumitem}

\usepackage{inconsolata}
\usepackage{listings}
\usepackage{graphicx}
\graphicspath{{../}{./}}

\usepackage{booktabs}

\title{StudyBench: Can Self-Evolution Squeeze Textbooks for Olympiad Capability?}

\author{
Yinghao Chen$^{1*}$,
Zixi Chen$^{1*}$,
Bingxiang He$^{1*\dagger}$,
Ziqing Qiao$^{1}$,
Huan-ang Gao$^{1}$, \\
\bfseries Yinuo Xu$^{1}$,
Yuxin Zuo$^{1}$,
Zeyuan Liu$^{1}$,
Yuhao Zhan$^{2}$,
Chaojun Xiao$^{1\dagger}$ \\[1ex]
{\normalfont\small $^{1}$Tsinghua University \quad $^{2}$Zhejiang University}
}

\begin{document}
\maketitle
{
  \renewcommand{\thefootnote}{\fnsymbol{footnote}}
  \footnotetext[1]{Equal contribution.}
  \footnotetext[2]{\raggedright Correspondence to: Bingxiang He \texttt{<hebx24@mails.tsinghua.edu.cn>}, Chaojun Xiao \texttt{<xcj@tsinghua.edu.cn>}.}
}
\begin{abstract}
  Humans need to study only a handful of well-written textbooks to master a discipline and attempt its hardest problems. We argue that an ideal self-evolution method should share the same property, that is \textbf{autonomously learning from raw training material for transferable problem-solving capability}. However, we still lack a direct measurement for it. We introduce \textbf{StudyBench}, a controlled physics benchmark that directly measures how efficiently a self-evolution method converts training material into capability. We organise the test set into an \textbf{Application Set}, consisting of difficult textbook problems and evaluating absorption ability, and a \textbf{Transfer Set}, consisting of olympiad-level problems and evaluating transfer ability. Benchmarking representative self-evolution methods across three base models, we find that improvements on the Application Set rarely translate to the harder Transfer Set. A guidance ablation exposes a \textbf{Guidance Gap}: even the strongest method closes only a small fraction of what the same material unlocks when supplied as in-context guidance. Besides, every method hits a \textbf{Compute Plateau}, saturating well before exhausting its compute budget. The remaining gap is therefore a method problem rather than a data or compute problem.   By offering a clean and controlled benchmark, StudyBench turns self-evolution progress from an open-ended pursuit into a measurable target for future research.
  Our code is released at \url{https://github.com/thunlp/StudyBench}.
\end{abstract}
    
\section{Introduction}

Self-evolution, the capacity for a model to keep improving on its
own without being capped by a limited supply of high-quality data,
shows the promise of true Artificial General Intelligence~\citep{goertzel2007artificial, silver2025welcome}. We argue
that any system worthy of that promise must succeed at two things at
once. First, it should continuously \emph{absorb} new knowledge from
its environment~\citep{yuan2026sebenchbenchmarkingselfevolutionknowledge}. Second, and more
critically, it should continuously \emph{evolve} absorbed
knowledge into transferable problem-solving capability, instead of merely memorizing or paraphrasing what it has seen~\citep{mirzadeh2025gsm, huang2025math, huang2026toolomnienablingopenworldtool}.
The second ability is arguably the harder of the two: absorption alone is bounded by what the training material already states; yet real-world problems seldom have direct precedents in the training material, and addressing them therefore demands \emph{transferable} problem-solving capability. Whether self-evolution can scale beyond the data it consumes
therefore turns on how efficiently it performs this
\textbf{knowledge-to-capability conversion}.

Despite a rapidly growing literature~\citep{gao2025survey, novikov2025alphaevolve, huang2025rzero} on self-evolution, we still lack a clear way to measure how effectively a method
performs this conversion. Static high-difficulty exams such as AIME and Humanity's Last Exam~\citep{phan2025humanity} only give a final score, conflating the algorithm's contribution with that of
the training data and the base model. Dynamic and lifelong-learning
benchmarks~\citep{castillo2024beyond,zheng2025lifelongagentbench,wan2025storybench,dou2026evalearn} test whether a model reuses earlier
experiences to do better on later problems; they address local adaptation, while we
operate at a higher level, testing conversion from experiences to transferable capability.
The deeper reason these existing evaluations fall short is that the
conversion itself is hard to measure directly: any such measurement
faces three obstacles. First, \textbf{the vanishing capability gap}: 
when the base model already passes the test set on its own, any 
post-training score merely surfaces existing capability rather than 
what the algorithm added. Second, \textbf{unreachable targets}: 
when the training material is insufficient for any algorithm to derive the test solutions. 
Third, \textbf{confounded attribution}: 
when the base model, the training material, and the algorithm vary 
together, thus the score conflates all three. 
To our knowledge, no existing evaluation eliminates all three at once.

We close this gap with \textbf{StudyBench}, aiming to directly reveal 
the enhancement brought by self-evolution methods. We
instantiate the setting in physics: 11 canonical
physics textbooks serve as the training material, factored into
three layers to fit requirements of different self-evolution methods: a \textbf{Corpus} of raw passages, \textbf{Instructions
with Answer}, and \textbf{Instructions without Answer}. 
The training material is paired with two
complementary test splits that form a built-in difficulty
progression. The \textbf{Application Set} consists of difficult
end-of-chapter exercises drawn from the textbooks themselves,
retained only when Qwen3-8B~\citep{yang2025qwen3technicalreport} does not solve them reliably;
it probes whether an algorithm has absorbed the training material.
The \textbf{Transfer Set} consists of olympiad-level theory
problems, retained only when Qwen3-8B alone fails but
succeeds with the help of textbook-grounded guidance, which is built from
passages taken directly from the training material; it probes whether the
method has additionally converted that material into capability
transferable to problems harder than any in the textbooks
themselves. We filter with Qwen3-8B because it sits at mid capability: strong enough that
a failed parent is a genuine gap, yet not so strong that the
retained set collapses or fails the guidance-based reachability
check.

This construction guarantees three properties by design:
\textbf{(1) Capability Gap}, that retained test problems lie
outside Qwen3-8B's reliable capability, so scores on that model
reflect ability gained rather than residual competence;
\textbf{(2) Reachability}, that every retained test problem is
solvable by utilizing the training material, eliminating unreachable targets; and
\textbf{(3) Controlled Attribution}, that the training material,
the test items, and the evaluation protocol are identical across
methods, so that within each base model the score isolates the
algorithm. Llama-3.2-3B-Instruct~\citep{grattafiori2024llama3herdmodels} and Opus~4.7~\citep{anthropic2025claude47} reuse the same
items, so cross-model comparison is on a shared problem set
rather than on per-model filters.

On StudyBench, we benchmark multiple representative self-evolution
methods across three base models and find that
\textbf{(i)} Application-Set gains remain local
to textbook exercises: on Qwen3-8B, GEPA lifts Application
$\mathrm{Par}@8$ from $17.05$ to $34.85$, yet Transfer
$\mathrm{Par}@8$ reaches only $7.04$; \textbf{(ii)} textbook-grounded guidance already certifies that every Transfer-Set parent is fully reachable from the training material, yet the strongest method raises $\mathrm{Par}@8$ only from $0.00$ to $7.04$ on Qwen3-8B; and \textbf{(iii)} on the self-evolution loops we profile, methods plateau well before exhausting their compute budget. The remaining gap is therefore a method problem rather than a data or compute problem, leaving substantial room for future work.

\section{StudyBench}
\label{sec:studybench}
\begin{figure*}[t]
  \centering
  \includegraphics[width=\textwidth]{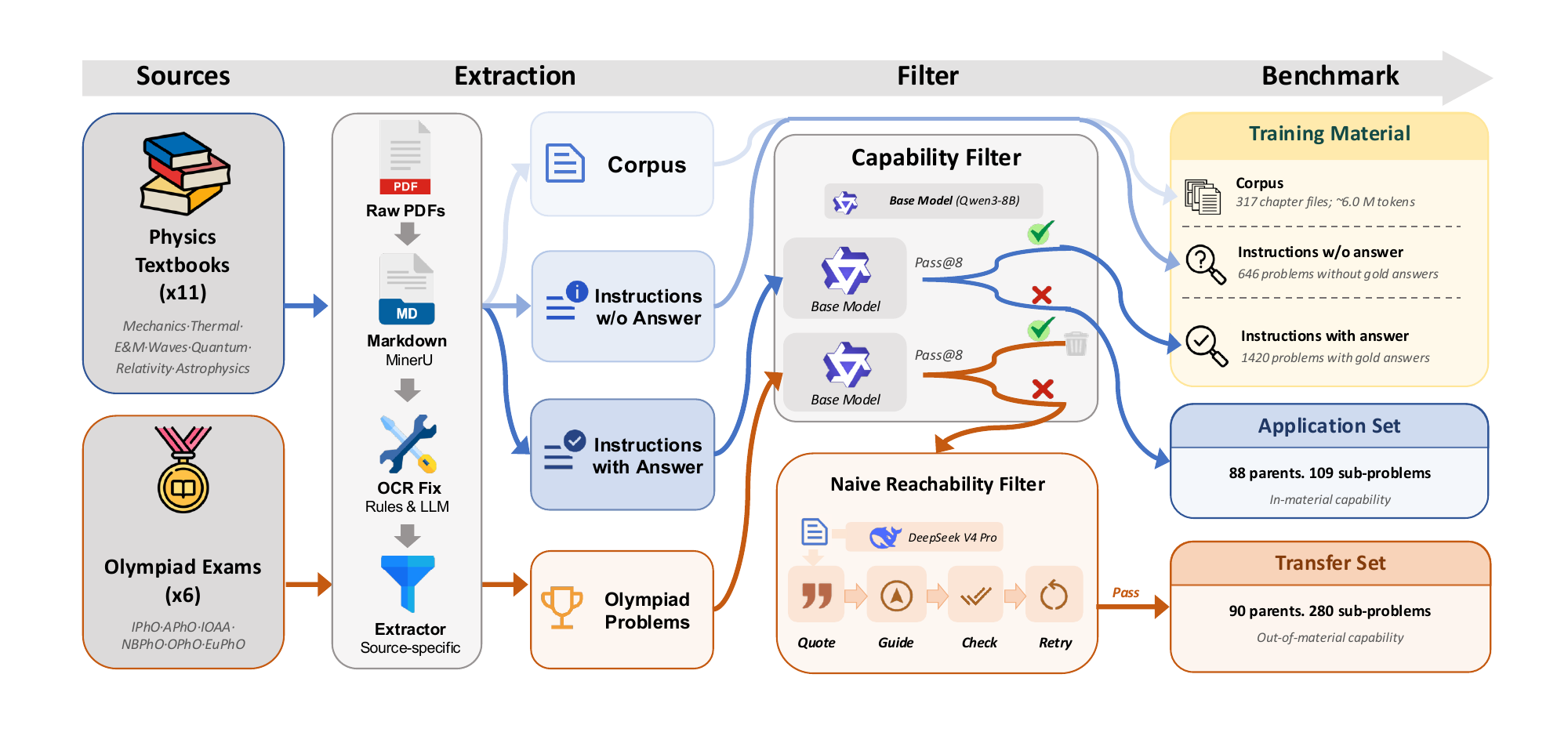}
  \caption{Construction pipeline for StudyBench. Textbooks are
  extracted into a \emph{Corpus}, \emph{Instructions with Answer}
  , and \emph{Instructions without Answer},
  which together form the training material; olympiad problems are
  extracted along a separate, test-only stream.   A \emph{Capability Filter} run on Qwen3-8B keeps items that
  model does not solve reliably; Llama-3.2-3B-Instruct and
  Opus~4.7 reuse the same items. Failed textbook problems become
  the \textbf{Application Set}; failed olympiad problems pass an
  additional \emph{Naive Reachability Filter} and become the
  \textbf{Transfer Set}.}
  \label{fig:overview}
\end{figure*}

StudyBench is inspired by the way humans master a discipline: a
motivated student typically needs no more than a handful of
well-chosen textbooks and some deliberate practice to attempt the
hardest problems in the field. A self-evolution method, placed in an analogous
setting, ought to demonstrate comparable capability. StudyBench
builds such a setting and provides a means to measure it. We
instantiate it in physics, where final answers are verifiable,
the set of standard textbooks is small and widely shared across physics curricula of top universities and olympiad training, and hard problems require both textbook knowledge and the reasoning capability to use it. These properties make physics an ideal setting for measuring capability rather than knowledge alone.

\subsection{Benchmark Construction}
\label{sec:construction}

Figure~\ref{fig:overview} summarises the construction. Each textbook yields three nested layers of training material:
the \textbf{Corpus} of raw passages contains the
\textbf{Instructions without Answer} (every exercise as a problem
statement), which in turn contains the \textbf{Instructions with
Answer} (the subset for which a gold answer is available from the
textbook or its official solution manual). Each
self-evolution method draws on whichever layer fits its training
paradigm. 

The test set, in turn, draws from two complementary sources of
escalating difficulty: difficult end-of-chapter problems from
these same textbooks (Table~\ref{tab:textbooks}) and problems from
six international physics and astronomy olympiads
(Table~\ref{tab:olympiads}). From both pools we keep only problems on which
Qwen3-8B does not succeed reliably, then subsample failed
textbook parents by sub-discipline so that no subject dominates;
failed competition parents all proceed. For the olympiad pool we
additionally retain only those Qwen3-8B can solve under a
textbook-grounded guidance trace, certifying that the answer is
reachable from the training material. Llama-3.2-3B-Instruct and
Opus~4.7 are scored on this same item set. The retained textbook problems form the
\textbf{Application Set}, which measures whether a method has
absorbed the training material well enough to apply it where it
was first introduced; the retained olympiad problems form the
\textbf{Transfer Set}, which measures whether the method has
additionally converted that material into capability that
transfers to problems harder than any in the textbooks
themselves. The remainder of this subsection details each step.

\vspace{+2pt}\noindent\textbf{Sources.}
The 11 textbooks (Table~\ref{tab:textbooks}) are each currently
adopted as a course text at top universities and independently
recommended either by olympiad-training coaches or on the
competitions' own preparation pages. By sub-discipline, the 11 textbooks jointly cover the syllabus of
all six olympiads.

\vspace{+2pt}\noindent\textbf{Extraction.}
All source materials are PDFs. We convert every PDF to Markdown via
MinerU~\citep{wang2026mineru2} and then fix common OCR errors (broken
super- and subscripts, malformed \LaTeX{}, dropped figure captions)
by first applying deterministic rules and then using an LLM to handle cases the rules cannot. With
Claude Opus 4.7 as a coding assistant, we then write a separate extractor for each textbook and each competition, since each source has its own layout. From each textbook we
extract its worked examples and end-of-chapter exercises, and from
each competition we extract every theory problem of every past
edition. For five textbooks we take exercise answers and reference
solutions from the matching official solution manual; for the rest they come from in-book answer
keys or end-of-chapter solutions (Table~\ref{tab:textbooks}).
Physics problems are commonly split into several
sub-problems sharing a common setup, so each extracted record stores
both the full problem statement and a list of per-sub-problem
entries carrying the sub-problem text, the reference solution where
given, and the gold answer. The gold answer of every sub-problem is
further classified by DeepSeek V4 Flash~\citep{deepseekai2026deepseekv4} into one of nine answer types: \texttt{NV} (numeric
value), \texttt{EX} (symbolic expression), \texttt{EQ} (equation),
\texttt{TUP} (ordered tuple), \texttt{IN} (interval), \texttt{MC}
(single multiple-choice letter), \texttt{TF} (boolean), \texttt{QL}
(short qualitative phrase), and \texttt{ALT} (alternative acceptable
forms of one answer); composite types (\texttt{TUP} and \texttt{ALT})
additionally carry a per-position sequence that tells the verifier
how to judge each slot. Appendix~\ref{app:extraction} gives the full
record schema and shows one example.

\vspace{+2pt}\noindent\textbf{Capability Filter.}
We run \textbf{Qwen3-8B} with $\mathrm{pass}@8$ sampling on both
pools and treat a parent as failed if no attempt solves every
sub-problem. Solved textbook problems stay in the training
material; solved competition problems are discarded. Every failed
competition parent advances to the Naive Reachability Filter.
Failed textbook parents do not: we subsample them by
sub-discipline so that hard subjects do not dominate the
Application Set, and send unselected failed parents back to the
training material. Subjects that would otherwise empty under a
strict zero-of-eight rule are kept in play by additionally
admitting $15$ parents that Qwen3-8B solved on exactly one of
eight attempts. These $15$ are the sole source of Qwen3-8B's
Application $\mathrm{Par}@8$ ($15/88=17.05\%$); the other $73$
Application parents, and every Transfer-Set parent, fail on all
eight attempts. Llama-3.2-3B-Instruct and Opus~4.7 reuse this
set. Appendix~\ref{app:discipline} reports the resulting
sub-discipline distribution. Since the prerequisite material lives in the
same chapter, the Application Set is reachable by construction.

\vspace{+2pt}\noindent\textbf{Naive Reachability Filter.}
Since we cannot precisely characterise the reachability boundary,
we adopt a naive but auditable proxy: a competition problem is
admitted only if Qwen3-8B can solve it under
teacher-distilled, textbook-grounded guidance. The procedure is
driven by a strong teacher DeepSeek V4 Pro~\citep{deepseekai2026deepseekv4} and
consists of five steps:
\begin{itemize}[topsep=0pt, partopsep=0pt, leftmargin=12pt, itemsep=-4.5pt]
  \item \textbf{Decompose.} For each sub-problem the teacher
    enumerates a minimal set of named knowledge points---concepts,
    laws or equations, techniques, and assumptions---required by
    the gold solution. Near-duplicate names are then
    canonicalised across the corpus.
  \item \textbf{Retrieve.} The training textbooks are first split
    into exposition and worked-example fragments. Each canonical
    knowledge point is matched to those fragments by a two-channel
    retriever: BM25 over normalised text and dense embeddings,
    fused by reciprocal rank fusion with a preference for
    in-domain books.
  \item \textbf{Verify.} The teacher scores every candidate on a
    $0$--$3$ coverage rubric and copies a short verbatim quote as
    evidence. A server-side check demotes any quote that does not
    actually appear in the fragment; only scores of $2$
    (applied/example) or $3$ (direct exposition) count as coverage.
  \item \textbf{Guide.} Given the verified passages and, for the
    teacher's own understanding, the gold solution, the teacher
    writes a methodological guidance that names which textbook
    concepts, formulae, and examples to use and in what order,
    without stating the answer or performing the key calculation.
    A dual leakage gate---deterministic redaction rules plus a
    separate teacher review---regenerates failing guidance up to
    twice and rule-sanitises the last attempt as a fallback.
  \item \textbf{Retry.} We admit the problem into the Transfer Set
    if Qwen3-8B solves it at least once in eight attempts
    under this grounded guidance.
\end{itemize}
Together these five steps form an explicit, textbook-auditable
witness that the gold answer is reachable by recombining the
training material. To check that this witness is not tied to one
teacher, we re-run the same pipeline with GLM-5.1 and evaluate
Qwen3-8B under the independently written traces: $56$ of $90$
Transfer-Set parents ($62.22$ $\mathrm{Par}@8$) and $242$ of
$280$ sub-problems ($86.43$ $\mathrm{Sub}@8$) remain solvable
(Appendix~\ref{app:alt-teacher}).

\vspace{+2pt}\noindent\textbf{Properties.}
By construction, the resulting benchmark satisfies three
properties. \textbf{(1) Capability Gap}, that retained test
problems lie outside Qwen3-8B's reliable capability;
\textbf{(2) Reachability}, that every retained test problem is
solvable by recombining content from the training material,
eliminating unreachable targets; and \textbf{(3) Controlled
Attribution}, that the training material, the test items, and the
evaluation protocol are fixed across methods, so that within each
base model a method's score reflects only what the method does
between them.

\subsection{Evaluation}
\label{sec:metrics}

\vspace{+2pt}\noindent\textbf{Protocol.}
Each method is free to draw on any subset of the training material to suit its training paradigm. We then
evaluate every evolved model on both test sets. Open-weight models
use $k{=}8$ samples per sub-problem; because of API cost, Opus~4.7
uses $k{=}1$. All runs share temperature $1.0$,
top-$p$ $0.95$, top-$k$ $20$, and a $32{,}768$-token cap. From
these samples we report two accuracies. Let $\mathcal{P}$ denote
the parent problems in a test set, $n_p$ the number of sub-problems of
parent $p$, and $c_{p,j,a} \in \{0,1\}$ the verifier's judgement on
the $j$-th sub-problem of parent $p$ in attempt $a$.
\textbf{Parent accuracy} ($\mathrm{Par}@k$) counts a parent correct
only if some single attempt solves every one of its sub-problems:
\[
\mathrm{Par}@k = \frac{1}{|\mathcal{P}|}\sum_{p \in \mathcal{P}}
\max_{1 \le a \le k} \prod_{j=1}^{n_p} c_{p,j,a}.
\]
\textbf{Sub-problem accuracy} ($\mathrm{Sub}@k$) flattens parents
into sub-problems and marks each correct if any of the $k$
attempts solves it:
\[
\mathrm{Sub}@k = \frac{1}{\sum_{p \in \mathcal{P}} n_p}
\sum_{p \in \mathcal{P}} \sum_{j=1}^{n_p}
\max_{1 \le a \le k} c_{p,j,a}.
\]
For open-weight models we repeat the $\mathrm{pass}@8$ evaluation
three times with independent sampling seeds and report the
mean $\pm$ standard deviation. For Opus~4.7 we report a single
$\mathrm{pass}@1$ run.

\vspace{+2pt}\noindent\textbf{Sub-problem Evaluation.}
Multi-part problems are scored sub-problem by sub-problem with
conversational continuity: at sub-problem $i$, the model sees the
shared stem, the $i-1$ prior sub-problem statements, and the
\emph{model's own} prior answers to them, but never the gold
solutions. When the model fails to produce a final answer for
sub-problem $i$, we insert a fixed placeholder stating the failure in the assistant slot
and continue with sub-problem $i+1$ rather than discarding the rest
of the parent. This isolates each sub-problem's correctness, so
$\mathrm{pass}@k$ can be reported per sub-problem and aggregated per
parent.

\vspace{+2pt}\noindent\textbf{Verifier.}
We build our verifier on the rule-based judger of
UG-Physics~\citep{xu2025ugphysics}, adding one new primitive answer type
\texttt{QL} (short qualitative phrases such as ``tidal forces'').
We also introduce a new field, \texttt{type\_sequence},
which lets the verifier dispatch composite answers (\texttt{TUP} and
\texttt{ALT}) onto position-wise primitive judgers. The resulting
verifier applies a different judging rule for every one of the nine
answer types. For leaderboard evaluation we additionally route failed
problems to \textbf{DeepSeek-V4-Flash-0731} as an LLM judger to enhance correctness. Appendix~\ref{app:verifier} reports a consistency analysis of the two-stage verifier. The
full judge prompt is reproduced in Appendix~\ref{app:judge}. When the verifier is wired into an RL-based
self-evolution method as a reward signal, we expose only this
rule-based part to avoid reward
hacking. 

\subsection{Contamination}
\label{sec:contamination}

The $11$ textbooks and the Olympiad archives are public, so a problem's statement or answer can, in principle, leak into a method's pipeline at two stages:
(a) the \emph{pretraining} of
the base model, or (b) \emph{training material} of a self-evolution method.
StudyBench neutralises both by design.

\vspace{+2pt}\noindent\textbf{Pretraining leakage is screened out by the Capability
Filter.} A problem enters either test set only if Qwen3-8B does
not solve it reliably under $\mathrm{pass}@8$ (zero successes, or,
for $15$ Application parents, a single success). Answers
Qwen3-8B has memorised well enough to recover consistently are
therefore removed from the test sets, regardless of whether the
problem statement appears verbatim in pretraining. The same items
are reused for Llama-3.2-3B-Instruct and Opus~4.7, so this screen
is defined with respect to Qwen3-8B.

\vspace{+2pt}\noindent\textbf{Training-material leakage is removed by redaction.}
Transfer Set problems come from olympiad theory exams, which do
not appear in any of the eleven textbooks that constitute the
training material. Application Set problems do come from those
textbooks, but for every retained parent we excise both the
problem statement and the reference solution (including any
back-of-book answer key or solution-manual entry) from the raw
markdown before assembling the training material, and we audit
the resulting corpus to confirm no verbatim residue remains;
Appendix~\ref{app:redaction} details the two-pass redaction
pipeline and the audit procedure. A method that memorises every
page of its training material therefore gains access to none of
the answers it will be asked for.

\subsection{Statistics}
\label{sec:stats}

\vspace{+2pt}\noindent\textbf{Test sets.}
The filters retain $88$ Application Set parents ($109$
sub-problems) and $90$ Transfer Set parents ($280$
sub-problems). The Application Set is subsampled so that no
sub-discipline dominates; the Transfer Set keeps every
competition parent that fails Qwen3-8B and then passes the
Naive Reachability Filter. Appendix~\ref{app:discipline} reports
both mixes. Together the two sets measure
\textbf{in-material} and \textbf{out-of-material} capability.

\vspace{+2pt}\noindent\textbf{Training material.}
Table~\ref{tab:trainstats} reports the per-layer counts of the
Corpus, the Instructions without Answer, and the Instructions
with Answer. The three layers cover the supervision regimes of the
major self-evolution families.

\begin{table}[h]
\centering
\small
\caption{Statistics of the three nested training-material
layers. \emph{Instructions with Answer} means we successfully extracted the instruction-answer pair from the textbook or its official solution manual, while \emph{Instructions without Answer} means we failed to find the corresponding answer.}
\label{tab:trainstats}
\begin{tabular}{lr}
\toprule
Layer & Count \\
\midrule
Corpus chapter files                       & 317 \\
~~total size (MB)                          & 18.19 \\
~~total tokens                             & $\sim 6.0$M \\
\midrule
Instructions without Answer                & 646 \\
Instructions with Answer                   & 1{,}420 \\
\bottomrule
\end{tabular}
\vspace{-4mm}
\end{table}

\begin{table*}[!t]
\centering
\small
\caption{Main results on StudyBench for open-weight models
($k{=}8$), grouped by training-material layer.
Qwen3-8B's Transfer $\mathrm{Par}@8$ is $0.00$ by construction.
$\Delta\mathrm{Sub}@8$ is relative to the same base model;
$\pm$ is over three independent $\mathrm{pass}@8$ evaluations.
All numbers in \%.}
\label{tab:main}
\begin{tabular}{lrrrrrr}
\toprule
 & \multicolumn{3}{c}{Application Set} & \multicolumn{3}{c}{Transfer Set} \\
\cmidrule(lr){2-4}\cmidrule(lr){5-7}
 & $\mathrm{Par}@8$ & $\mathrm{Sub}@8$ & $\Delta\mathrm{Sub}@8$ & $\mathrm{Par}@8$ & $\mathrm{Sub}@8$ & $\Delta\mathrm{Sub}@8$ \\
\midrule
Qwen3-8B           & 17.05 $\pm$ 1.14 & 29.36 $\pm$ 2.75 & -- & 0.00 & 56.43 $\pm$ 1.89 & -- \\
\midrule
\multicolumn{7}{l}{\emph{Corpus}} \\
~~Bonito~\citep{nayak2024learning}     & 21.21 $\pm$ 3.47 & 28.13 $\pm$ 3.47 & $-1.23$ & 4.44 $\pm$ 1.11 & 35.00 $\pm$ 0.36 & $-21.43$ \\
\multicolumn{7}{l}{\emph{Instructions with Answer}} \\
~~GRPO~\citep{shao2024deepseekmathpushinglimitsmathematical} & 28.41 $\pm$ 1.97 & 38.84 $\pm$ 1.40 & $+9.48$ & 4.07 $\pm$ 1.70 & 57.02 $\pm$ 1.09 & $+0.59$ \\
~~GEPA~\citep{agrawal2025gepa}       & \textbf{34.85 $\pm$ 1.31} & \textbf{44.34 $\pm$ 1.40} & $\mathbf{+14.98}$ & \textbf{7.04 $\pm$ 2.80} & \textbf{58.57 $\pm$ 1.43} & $\mathbf{+2.14}$ \\
~~ACE~\citep{zhang2025agentic}           & 31.06 $\pm$ 1.74 & 41.90 $\pm$ 1.06 & $+12.54$ & 2.96 $\pm$ 2.31 & 57.26 $\pm$ 1.09 & $+0.83$ \\
\multicolumn{7}{l}{\emph{Instructions without Answer}} \\
~~TTRL~\citep{zuo2025ttrltesttimereinforcementlearning} & 28.79 $\pm$ 3.65 & 38.84 $\pm$ 2.80 & $+9.48$ & 2.59 $\pm$ 1.70 & \textbf{58.57 $\pm$ 1.86} & $\mathbf{+2.14}$ \\
~~Intuitor~\citep{zhao2025intuitor}  & 26.89 $\pm$ 2.37 & 38.23 $\pm$ 2.65 & $+8.87$ & 5.56 $\pm$ 1.92 & 58.21 $\pm$ 0.36 & $+1.78$ \\
\multicolumn{7}{l}{\emph{Data-free}} \\
~~R-Zero~\citep{huang2025rzero}      & 29.55 $\pm$ 5.90 & 39.14 $\pm$ 5.30 & $+9.78$ & 4.07 $\pm$ 2.80 & 58.10 $\pm$ 1.03 & $+1.67$ \\
\multicolumn{7}{l}{\emph{Guided (Corpus)}} \\
~~Naive Guidance & -- & -- & -- & $100.00$ & $100.00$ & $+43.57$ \\
\midrule
Llama-3.2-3B-Instruct           & 9.47 $\pm$ 0.66 & 14.37 $\pm$ 0.53 & -- & $0.00$ & $14.88\pm1.76$ & -- \\
\midrule
\multicolumn{7}{l}{\emph{Corpus}} \\
~~Bonito~\citep{nayak2024learning}     & 10.98 $\pm$ 2.62 & 15.90 $\pm$ 2.31 & $+1.53$ & 1.85 $\pm$ 1.28 & 17.86 $\pm$ 1.86 & $+2.98$ \\
\multicolumn{7}{l}{\emph{Instructions with Answer}} \\
~~GRPO~\citep{shao2024deepseekmathpushinglimitsmathematical} & 10.98 $\pm$ 1.31 & 14.07 $\pm$ 1.40 & $-0.30$ & \textbf{2.59 $\pm$ 1.28} & \textbf{22.74 $\pm$ 1.44} & $\mathbf{+7.86}$ \\
~~GEPA~\citep{agrawal2025gepa}       & 11.74 $\pm$ 1.31 & 14.98 $\pm$ 1.06 & $+0.61$ & 2.22 $\pm$ 2.94 & 17.26 $\pm$ 1.25 & $+2.38$ \\
~~ACE~\citep{zhang2025agentic}           & 9.85 $\pm$ 3.65 & 14.37 $\pm$ 2.95 & $+0.00$ & $1.85\pm0.02$ & $16.79\pm1.29$ & $+1.91$ \\
\multicolumn{7}{l}{\emph{Instructions without Answer}} \\
~~TTRL~\citep{zuo2025ttrltesttimereinforcementlearning} & 9.85 $\pm$ 0.66 & 12.85 $\pm$ 0.92 & $-1.52$ & \textbf{2.59 $\pm$ 0.64} & 19.76 $\pm$ 1.49 & $+4.88$ \\
~~Intuitor~\citep{zhao2025intuitor}  & \textbf{12.50 $\pm$ 2.27} & \textbf{16.82 $\pm$ 1.91} & $\mathbf{+2.45}$ & 0.74 $\pm$ 0.64 & 20.60 $\pm$ 0.82 & $+5.72$ \\
\multicolumn{7}{l}{\emph{Data-free}} \\
~~R-Zero~\citep{huang2025rzero}      & 5.68 $\pm$ 1.14 & 10.70 $\pm$ 1.06 & $-3.67$ & 0.74 $\pm$ 0.64  & 18.69 $\pm$ 1.15 & $+3.81$ \\
\multicolumn{7}{l}{\emph{Guided (Corpus)}} \\
~~Naive Guidance & -- & -- & -- & 10.74 $\pm$ 3.39 & 46.79 $\pm$ 0.36 & $+31.91$ \\
\bottomrule
\end{tabular}
\end{table*}

\begin{table*}[!t]
\centering
\small
\caption{Results on StudyBench for Opus~4.7 with Claude Code
($k{=}1$), on the same item set as Table~\ref{tab:main}.
A single $\mathrm{pass}@1$ run. All numbers in \%.}
\label{tab:opus}
\begin{tabular}{lrrrrrr}
\toprule
 & \multicolumn{3}{c}{Application Set} & \multicolumn{3}{c}{Transfer Set} \\
\cmidrule(lr){2-4}\cmidrule(lr){5-7}
 & $\mathrm{Par}@1$ & $\mathrm{Sub}@1$ & $\Delta\mathrm{Sub}@1$ & $\mathrm{Par}@1$ & $\mathrm{Sub}@1$ & $\Delta\mathrm{Sub}@1$ \\
\midrule
Opus 4.7 with Claude Code           & 59.09 & 66.06 & -- & 40.00 & 72.14 & -- \\
\midrule
\multicolumn{7}{l}{\emph{Instructions with Answer}} \\
~~GEPA~\citep{agrawal2025gepa}       & 51.14 & 57.80 & $-8.26$ & 41.11 & 72.14 & $+0.00$ \\
~~ACE~\citep{zhang2025agentic}           & \textbf{60.23} & \textbf{63.30} & $-2.76$ & 33.33 & 66.79 & $-5.35$ \\
~~EvoSkill~\citep{alzubi2026evoskillautomatedskilldiscovery}  & 52.27 & 58.72 & $-7.34$ & \textbf{43.33} & \textbf{73.57} & $\mathbf{+1.43}$ \\
\multicolumn{7}{l}{\emph{Guided (Corpus)}} \\
~~Naive Guidance & -- & -- & -- & 63.33 & 84.64 & $+12.50$ \\
\bottomrule
\end{tabular}
\end{table*}

\section{Experiments}
\label{sec:experiments}

\vspace{+2pt}\noindent\textbf{Baselines.}
We benchmark a representative set of self-evolution methods grouped by which layer of the training
material each one consumes.
\textbf{(1) Corpus.} \textbf{Bonito}~\citep{nayak2024learning} runs
a task-conditioned generator over textbook passages to synthesise
question--answer pairs and performs supervised fine-tuning on the
base model with them. \textbf{Naive Guidance} is not a training
method: it injects textbook-grounded traces built from the same
Corpus at inference time, and serves as the reachability ceiling
on the Transfer Set.
\textbf{(2) Instructions with Answer.}
\textbf{GRPO}~\citep{shao2024deepseekmathpushinglimitsmathematical}
is a supervised RL reference: it trains on these labelled problems
with a gold outcome reward. \textbf{GEPA}~\citep{agrawal2025gepa}
treats them as a development set and evolves a system prompt via
reflective genetic search, while \textbf{ACE}~\citep{zhang2025agentic}
distils them into an in-context playbook of formulae, strategies,
and common pitfalls; both artefacts are injected into the system
message at inference time without any weight update.
\textbf{(3) Instructions without Answer.}
\textbf{TTRL}~\citep{zuo2025ttrltesttimereinforcementlearning}
and \textbf{Intuitor}~\citep{zhao2025intuitor} both run label-free
RL on these problems, rewarding majority-vote consistency and the
policy's own self-certainty, respectively.
\textbf{(4) Data-free.} \textbf{R-Zero}~\citep{huang2025rzero}
bootstraps capability from Challenger-Solver self-play under
self-consistency rewards. We include it to see how well a method
that uses no training data at all can do on StudyBench.

\vspace{+2pt}\noindent\textbf{Setup.}
The test set is filtered with \textbf{Qwen3-8B}~\citep{yang2025qwen3technicalreport} (thinking mode
enabled). We then evaluate the same items on three base models of
increasing capability: Llama-3.2-3B-Instruct~\citep{grattafiori2024llama3herdmodels}, Qwen3-8B, and
Opus~4.7~\citep{anthropic2025claude47} with Claude Code. For each method we reproduce from its
official repository and make only minimal modifications to fit
StudyBench's training material and evaluation protocol. All
open-weight self-evolution methods run on a single
$8{\times}$NVIDIA-A800-80GB node.
Appendix~\ref{app:budgets} lists per-method modifications and
full configurations.

\vspace{+2pt}\noindent\textbf{Results.}
Tables~\ref{tab:main} and~\ref{tab:opus} report the main results. Because the
item set is filtered with Qwen3-8B, that model's Transfer
$\mathrm{Par}@8$ is $0.00$ by construction, while its
Application $\mathrm{Par}@8$ of $17.05$ is the $15$ coverage
parents admitted at one-of-eight; Llama-3.2-3B-Instruct and
Opus~4.7 are not guaranteed a zero parent score on the same
items. On Qwen3-8B, most methods absorb the textbooks:
Application $\mathrm{Sub}@8$ rises by $+8.87$ to $+14.98$, with
\textbf{GEPA} leading on both Application metrics
($34.85$ $\mathrm{Par}@8$, $+14.98$ $\Delta\mathrm{Sub}@8$).
Those gains stay local. Transfer $\mathrm{Sub}@8$ moves by at
most $+2.14$ (GEPA and TTRL), and the best Transfer
$\mathrm{Par}@8$ is only $7.04$ against a $100\%$
Corpus-grounded guidance ceiling.
\textbf{Bonito} is the exception: synthetic-data SFT erodes
Qwen3-8B's thinking behaviour, so problems the base model
already solved become unsolvable ($-1.23$ Application and
$-21.43$ Transfer $\Delta\mathrm{Sub}@8$).
The other two models on the same items confirm the pattern.
Llama-3.2-3B-Instruct barely moves on Application (peak
$+2.45$ under Intuitor), while supervised GRPO leads Transfer
($+7.86$); Opus~4.7 already solves much of the set, and
context-evolution methods mostly regress on Application.
Section~\ref{sec:rq1} attributes the remaining Transfer
headroom to the \textbf{Guidance Gap}---the accuracy a method
recovers when textbook-grounded guidance is supplied at
inference but not internalised by training.

\section{Analysis}
\label{sec:analysis}

We organise our analysis around two research questions: \textbf{RQ1:} \emph{how much of the corpus-reachable knowledge actually migrates into standalone transfer capability?} \textbf{RQ2:} \emph{can the remaining gap be closed by simply running each self-evolution loop longer?}

\subsection{The Guidance Gap (RQ1)}
\label{sec:rq1}

The Application Set probes whether a method has \emph{absorbed}
the training material; the Transfer Set probes whether that
material has \emph{transferred} to problems strictly harder than
any in the textbooks. The two probes diverge in
Table~\ref{tab:main}. On Qwen3-8B, Application
$\Delta\mathrm{Sub}@8$ ranges from $+8.87$ to $+14.98$ for
every method except Bonito, yet Transfer
$\Delta\mathrm{Sub}@8$ is at most $+2.14$ and Transfer
$\mathrm{Par}@8$ stays in the single digits. \textbf{RQ1}
asks how much of the corpus-reachable knowledge certified by
our Naive Reachability Filter has actually migrated into
standalone transfer capability, rather than remaining merely
reachable.

\vspace{+2pt}\noindent\textbf{Guidance ablation.}
To answer this, we re-run the Transfer Set under the
textbook-grounded guidance produced during construction
(\S\ref{sec:construction}), injecting them at inference.
Table~\ref{tab:guidance} reports the ablation for Qwen3-8B and
the five self-evolution methods we re-ran under guidance; We define the \textbf{Guidance Gap}
as the accuracy a method recovers when grounded guidance is
supplied at inference but not internalised by training:
\begin{equation}
  \text{Guidance Gap} =
  \text{Acc}_{\text{guided}} - \text{Acc}_{\text{solo}},
\end{equation}
where $\text{Acc}_{\text{solo}}$ is Transfer-Set accuracy
under the solo protocol of \S\ref{sec:metrics} and
$\text{Acc}_{\text{guided}}$ uses the same Naive Reachability
Filter traces in the system prompt. Both quantities are
reported for $\mathrm{Par}@8$ and $\mathrm{Sub}@8$. Closing
the gap means the method has converted into standalone
capability what, for Qwen3-8B, only an in-context guidance
trace could surface.

\begin{table}[t]
\centering
\small
\setlength{\tabcolsep}{3.5pt}
\caption{Transfer-Set accuracy with textbook-grounded guidance
on Qwen3-8B. Solo accuracies repeat Table~\ref{tab:main};
guided accuracies inject the Naive Reachability Filter traces
at inference. \textbf{Guidance Gap} $=
\text{Acc}_{\text{guided}} - \text{Acc}_{\text{solo}}$.
Qwen3-8B's guided scores are $100$ by construction of that
filter. Appendix~\ref{app:alt-teacher} reports the same protocol
with independently written GLM-5.1 traces.}
\label{tab:guidance}
\begin{tabular}{lrrrrrr}
\toprule
 & \multicolumn{3}{c}{$\mathrm{Par}@8$} & \multicolumn{3}{c}{$\mathrm{Sub}@8$} \\
\cmidrule(lr){2-4}\cmidrule(lr){5-7}
Method & Solo & Guided & Gap & Solo & Guided & Gap \\
\midrule
Qwen3-8B  & 0.00 & 100.00 & 100.00 & 56.43 & 100.00 & 43.57 \\
Bonito    & 4.44 & 23.33 & 18.89 & 35.00 & 68.93 & 33.93 \\
GEPA      & 7.04 & 90.00 & 82.96 & 58.57 & 97.14 & 38.57 \\
ACE    & 2.96 & 88.89 & 85.93 & 57.26 & 97.14 & 39.88 \\
Intuitor  & 5.56 & 90.00 & 84.44  & 58.21 & 98.21 & 40.00 \\
R-Zero    & 4.07 & 90.00 & 85.93 & 58.10 & 98.21 & 40.11 \\
\bottomrule
\end{tabular}
\end{table}

\vspace{+2pt}\noindent\textbf{A high, method-stable ceiling---except under SFT.}
By construction of the Reachability Filter, Qwen3-8B under
guidance solves every Transfer parent ($100$ $\mathrm{Par}@8$
and $\mathrm{Sub}@8$). The same traces remain highly effective
after context evolution or label-free RL: GEPA, ACE, Intuitor,
and R-Zero all sit at $88.89$--$90.00$ guided
$\mathrm{Par}@8$ and $97.14$--$98.21$ guided
$\mathrm{Sub}@8$. Training therefore barely moves the
ceiling. \textbf{Bonito} is the outlier: after synthetic-data
SFT its guided $\mathrm{Par}@8$ falls to $23.33$ and guided
$\mathrm{Sub}@8$ to $68.93$. The same collapse of thinking
behaviour that hurts solo accuracy also prevents the model
from following guidance that the untrained backbone can use.

\vspace{+2pt}\noindent\textbf{Solo gains close almost none of the gap.}
Relative to Qwen3-8B's original headroom ($100$
$\mathrm{Par}@8$ points; $43.57$ $\mathrm{Sub}@8$ points), the
strongest solo run is GEPA at $7.04$ $\mathrm{Par}@8$ and
$+2.14$ $\Delta\mathrm{Sub}@8$---closing $\mathbf{7\%}$ of
the parent gap and $\mathbf{5\%}$ of the sub-problem gap.
Intuitor ($5.56$ / $+1.78$) and R-Zero ($4.07$ / $+1.67$)
close even less; ACE trails on parents ($2.96$). GRPO and
TTRL, which we did not re-run under guidance, have solo
Transfer scores in the same band ($4.07$ / $+0.59$ and
$2.59$ / $+2.14$). Bonito's solo Transfer $\mathrm{Sub}@8$
falls below the base model, so it widens rather than closes
the sub-problem gap. The Application Set, where the same
methods (Bonito aside) gain $+8.87$ to $+14.98$
$\mathrm{Sub}@8$, cannot see this failure: absorption and
transfer remain complementary probes.

\vspace{+2pt}\noindent\textbf{Substantial headroom remains.}
Because guidance still unlocks about $90\%$ of Transfer
parents for every method except Bonito, what remains is an
\emph{internalisation} problem rather than a
\emph{reachability} problem: the textbooks already contain
what is needed to solve the retained olympiad parents, yet no
method converts more than a sliver of that into standalone
capability. Llama-3.2-3B-Instruct and Opus~4.7 reuse the same
items and the same traces (Tables~\ref{tab:main} and~\ref{tab:opus}); their
guidance ceilings are lower ($10.74$ and $63.33$
$\mathrm{Par}@8$) because the filter was built for Qwen3-8B,
but the qualitative gap is unchanged---the strongest solo
run still leaves most of the available headroom on the table.

\subsection{The Compute Plateau (RQ2)}
\label{sec:rq2}

RQ1 left most of the Guidance Gap unclosed: even GEPA, the strongest method, recovers only $7\%$ of Qwen3-8B's parent-level Transfer headroom. \textbf{RQ2} asks whether the remaining gap can be closed by simply running each self-evolution loop longer.

\vspace{+2pt}\noindent\textbf{Setup.}
For each of the five self-evolution loops we profile on
Qwen3-8B, we evaluate the model checkpoint (Bonito, Intuitor,
R-Zero) or inference-time artefact (GEPA, ACE) at intermediate
snapshots and plot Application-Set $\mathrm{Sub}@8$ against
cumulative GPU-time on NVIDIA A800 GPUs. Compute scales
differ by nearly two orders of magnitude---from $8.12$
GPU hour for Bonito to $614$ GPU hour for R-Zero---so the
curves cannot share an axis. We show \textbf{ACE} in the main
text (Figure~\ref{fig:ace-curve}) because we ran it under three
random seeds, and place the other four curves in
Appendix~\ref{app:curves}. 

\begin{figure}[t]
  \centering
  \includegraphics[width=\columnwidth]{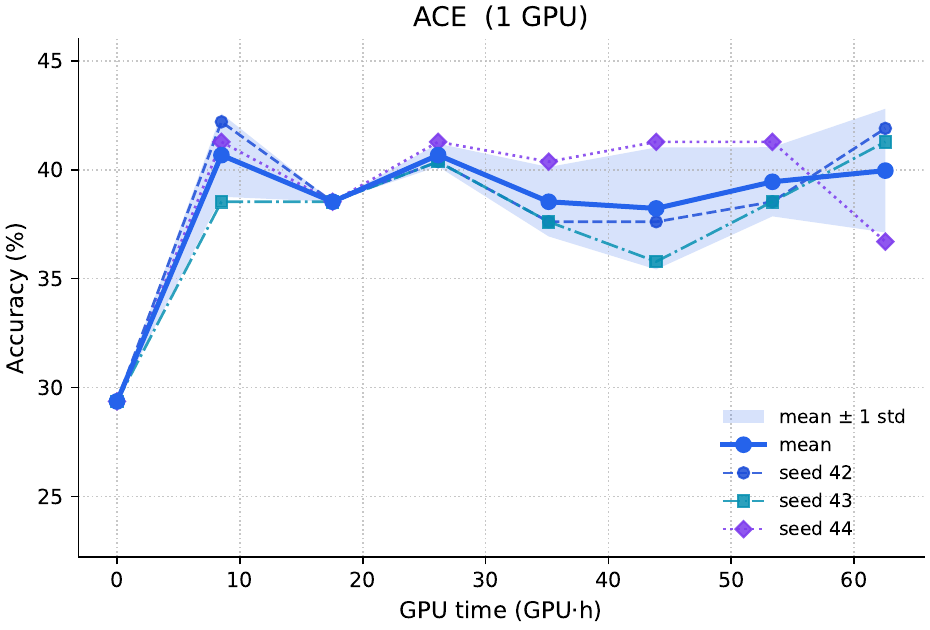}
  \caption{Application-Set $\mathrm{Sub}@8$ of \textbf{ACE}
  against cumulative GPU-time (1$\times$A800). The solid line is
  the mean of three seeds; the band is
  $\pm 1$ standard deviation. Accuracy jumps from the Qwen3-8B
  baseline ($\approx 29.4\%$) to $\approx 40\%$ by the
  $8.50$ GPU hour snapshot, then the mean stays in a
  $38$--$41\%$ band through the $62.54$ GPU hour endpoint. Late
  seed-to-seed scatter does not resume a climb. The same
  early-plateau pattern holds for the other four methods
  (Appendix~\ref{app:curves}).}
  \label{fig:ace-curve}
\end{figure}

\vspace{+2pt}\noindent\textbf{The plateau is not a single-run artefact.}
ACE's three seeds all make the same early jump and then stop
improving: after the $8.50$ GPU hour snapshot the mean never
leaves a narrow band, even though the remaining $54.0$
GPU hour are most of the $62.54$ GPU hour budget. Seed $44$
drops at the $62.54$ GPU hour snapshot and seed $42$ recovers;
neither trajectory looks like continued learning. The
table's ACE $\mathrm{Sub}@8$ of $41.90$ sits inside this
band.

The other four loops (Appendix~\ref{app:curves}) saturate in
the same way: GEPA climbs then sits flat; Intuitor and
R-Zero peak and decline; Bonito never leaves the baseline
band. In every case the late phase is a noisy plateau or a
decline, not a second climb.

\vspace{+2pt}\noindent\textbf{Compute is not the bottleneck.}
The remaining headroom from RQ1 therefore cannot be closed by
running any of these loops longer. Extra GPU-time past
saturation moves Application accuracy by at most a few noisy
points, even though total compute spans $76\times$
($8.12$ to $614$ GPU hour) and the highest plateau (GEPA's
$44.3\%$) is not the most expensive. Closing more of the gap
will require a different self-evolution loop, not a longer
one.

\section{Related Work}
\label{sec:related}

\subsection{Self-Evolving Methods}

Self-evolution methods differ mainly in where the improvement signal
comes from. Self-play and intrinsic-reward approaches such as
R-Zero~\citep{huang2025rzero}, Absolute Zero~\citep{zhao2026absolute},
SPICE~\citep{liu2025spice}, CPMobius~\citep{li2026cpmobiusiterativecoachplayerreasoning},
and Intuitor~\citep{zhao2025intuitor} generate or score new practice
without relying on a fixed external answer set. Context-level methods
such as GEPA~\citep{agrawal2025gepa} and
ACE~\citep{zhang2025agentic} leave model weights unchanged and instead
evolve prompts or in-context playbooks. SEAL~\citep{zweiger2026self}
occupies a third point in the design space: it trains a model to emit
self-edits that transform passages into finetuning data.

A related line studies improvement through memory, reflection, or
test-time search. Voyager~\citep{wang2023voyager},
Reflexion~\citep{shinn2023reflexion}, ExpEL~\citep{zhao2024expel}, and
ReasoningBank~\citep{ouyang2025reasoningbank} accumulate reusable
experience across trajectories, while
AlphaEvolve~\citep{novikov2025alphaevolve} applies evolutionary search
at inference time. These systems demonstrate many ways to make a model
adapt after deployment, but they are usually evaluated with implicit or
freely chosen training data. As a result, it is difficult to attribute
gains to a specific corpus or compare methods from different families
on equal footing.

\subsection{Benchmarks for Self-Evolution}

Dynamic and lifelong-learning benchmarks such as the LTM
Benchmark~\citep{castillo2024beyond},
LifelongAgentBench~\citep{zheng2025lifelongagentbench},
StoryBench~\citep{wan2025storybench}, and
EvaLearn~\citep{dou2026evalearn} measure adaptation over interaction
streams, but typically leave the training material implicit or
submitter-chosen. More targeted benchmarks, including
SE-Bench~\citep{yuan2026sebenchbenchmarkingselfevolutionknowledge},
NewtonBench~\citep{zheng2025newtonbench}, and
Frontier-Eng~\citep{chi2026frontier}, study knowledge internalisation,
scientific-law discovery, or engineering agents.

These benchmarks are closest to StudyBench in spirit, but their units
of comparison differ from ours. Some emphasise local adaptation within
an interaction stream; others evaluate open-ended agent improvement on
tasks whose useful training evidence is not fixed in advance.
StudyBench is complementary: it fixes the source corpus, enforces both
a capability gap and reachability, and provides a guidance ceiling for
measuring how much corpus-reachable capability a method internalises.

\section{Conclusion}
\label{sec:conclusion}

We introduce \textbf{StudyBench}, a controlled physics
benchmark that measures how efficiently a self-evolution
method converts a fixed corpus into transferable
problem-solving capability. Pairing 11 textbooks with an
Application Set of unsolved textbook exercises and a Transfer
Set of olympiad problems certified reachable under
textbook-grounded guidance, it isolates
knowledge-to-capability conversion from a vanishing
capability gap, unreachable targets, and confounded
attribution. Application-Set gains do not become olympiad
capability: on Qwen3-8B, GEPA lifts Application
$\mathrm{Par}@8$ from $17.05$ to $34.85$, yet Transfer
$\mathrm{Par}@8$ reaches only $7.04$ against a $100\%$
guidance ceiling, and the same local-gain pattern holds on
Llama-3.2-3B-Instruct and Opus~4.7. The profiled loops
additionally hit a Compute Plateau. The remaining gap is
therefore a method problem rather than a data or compute
problem---a distinction StudyBench makes directly measurable
for future research.

\section*{Acknowledgements}
This work is supported by the China National Postdoctoral Program for Innovative Talents (grant no. BX20250388), Tsinghua University (Department of Computer Science and Technology)-Siemens Ltd., China Joint Research Center for Industrial Intelligence and Internet of Things (JCIIOT), Institute Guo Qiang at Tsinghua University.

\section*{Limitations}
\label{sec:limitations}

The test set is filtered with Qwen3-8B, then reused for
Llama-3.2-3B-Instruct and Opus~4.7; a capability gap and a
$100\%$ guidance ceiling are therefore guaranteed only for
Qwen3-8B under the DeepSeek V4 Pro traces, and Opus~4.7 already
solves a substantial fraction of both splits. Independently
written GLM-5.1 traces recover $62.22$ $\mathrm{Par}@8$ on the
same Transfer-Set items (Appendix~\ref{app:alt-teacher}).
The setting is instantiated in physics over 11
textbooks; while the construction principles---Capability
Filter, Naive Reachability Filter, and two-level test
design---are domain-agnostic, we have not verified that they
replicate in other disciplines. On the Application Set we
additionally admit $15$ parents that Qwen3-8B solved once in
eight attempts, a limited relaxation of the Capability Gap in
order to keep easier subjects represented. Compute constraints
additionally limit the guidance ablation
(Table~\ref{tab:guidance}) and the compute-plateau curves; we
do not claim either result for every method--model pair.
Verifier consistency (Appendix~\ref{app:verifier}) is measured
on Qwen3-8B Transfer-Set rollouts only.

\bibliography{custom}

\clearpage
\appendix

\section{Source materials}
\label{app:sources}

Table~\ref{tab:textbooks} lists the eleven physics textbooks that
make up the training material $\mathcal{C}$, alongside the
sub-discipline each one covers. Table~\ref{tab:olympiads} lists the
six international physics and astronomy olympiads that contribute to
the competition pool; for each contest we collected every past
edition.
Together the two tables make explicit the syllabus-level match
between $\mathcal{C}$ and the test pool: every sub-discipline
labelled in Table~\ref{tab:textbooks} is also a sub-discipline
tested by at least one olympiad in Table~\ref{tab:olympiads}.

\begin{table}[h]
\centering
\footnotesize
\setlength{\tabcolsep}{4pt}
\caption{The 11 physics textbooks comprising StudyBench's training
material. Textbooks marked \textdagger{} are paired with an official
solution manual, which we bundle alongside the textbook and use
to extract exercise answers.}
\label{tab:textbooks}
\begin{tabular}{ll}
\toprule
Textbook & Sub-discipline \\
\midrule
\emph{Introduction to Classical Mechanics} & Mechanics \\
\emph{Introduction to Mechanics}\textdagger & Mechanics \\
\addlinespace[2pt]
\emph{Electricity and Magnetism}\textdagger & Electromagnetism \\
\emph{Introduction to Electrodynamics}\textdagger & Electromagnetism \\
\addlinespace[2pt]
\emph{Concepts in Thermal Physics}\textdagger & Thermal Physics \\
\addlinespace[2pt]
\emph{The Physics of Waves}\textdagger & Waves \\
\addlinespace[2pt]
\emph{Quantum Physics} & Quantum Physics \\
\addlinespace[2pt]
\emph{Special Relativity} & Relativity \\
\emph{Spacetime Physics} & Relativity \\
\addlinespace[2pt]
\shortstack[l]{\emph{An Introduction to}\\ \emph{Modern Astrophysics}} & Astrophysics \\
\emph{Schaum's Outline of Astronomy} & Astrophysics \\
\bottomrule
\end{tabular}
\end{table}

\begin{table}[h]
\centering
\footnotesize
\setlength{\tabcolsep}{4pt}
\caption{The six international physics olympiads we utilized.}
\label{tab:olympiads}
\begin{tabular}{ll}
\toprule
Acronym & Full name \\
\midrule
APhO  & Asian Physics Olympiad \\
\addlinespace[2pt]
EuPhO & European Physics Olympiad \\
\addlinespace[2pt]
IPhO  & International Physics Olympiad \\
\addlinespace[2pt]
IOAA  & Intl.\ Olympiad on Astronomy \& Astrophysics \\
\addlinespace[2pt]
NBPhO & Nordic-Baltic Physics Olympiad \\
\addlinespace[2pt]
OPhO  & Online Physics Olympiad (Invitational) \\
\bottomrule
\end{tabular}
\end{table}

\section{Distribution of the training material}
\label{app:release}

The eleven textbooks in Table~\ref{tab:textbooks} are
in-copyright commercial works. We therefore do not release
their PDFs, nor the \textbf{Corpus} of raw passages parsed
from them. The public release contains only the instruction
layers that we extract and process---\textbf{Instructions
with Answer} and \textbf{Instructions without
Answer}---together with the scripts that rebuild the Corpus
from a reader's own legal copies of the books. Methods that
require the Corpus layer must run that pipeline locally;
methods that consume only the instruction layers can use the
released files directly.
The extracted textbook instructions and the competition
problems are released for academic research only and must
not be used for commercial purposes.

\section{Sub-discipline mix of the two test sets}
\label{app:discipline}

Table~\ref{tab:discipline} reports the parent and sub-problem
counts of Application Set and Transfer Set by sub-discipline. 

\begin{table}[h]
\centering
\small
\caption{Sub-discipline mix of the Application Set ($88$ parents,
$109$ sub-problems) and the Transfer Set ($90$ parents, $280$
sub-problems).}
\label{tab:discipline}
\begin{tabular}{lrrrr}
\toprule
 & \multicolumn{2}{c}{Application Set} & \multicolumn{2}{c}{Transfer Set} \\
\cmidrule(lr){2-3}\cmidrule(lr){4-5}
Sub-discipline & Parents & Subs & Parents & Subs \\
\midrule
Mechanics          & 18 & 22 & 28 & 89 \\
Electromagnetism   & 19 & 23 & 18 & 59 \\
Astrophysics       & 14 & 18 & 23 & 67 \\
Quantum Physics    & 15 & 17 &  4 & 15 \\
Thermal Physics    & 12 & 15 & 12 & 31 \\
Relativity         &  3 &  3 &  2 &  7 \\
Waves              &  7 & 11 &  3 & 12 \\
\midrule
\textbf{Total}     & \textbf{88} & \textbf{109} & \textbf{90} & \textbf{280} \\
\bottomrule
\end{tabular}
\end{table}

\section{Alternative-teacher guidance}
\label{app:alt-teacher}

The Naive Reachability Filter (\S\ref{sec:construction}) is
driven by DeepSeek V4 Pro. To test whether the retained Transfer
Set is reachable only under that teacher's traces, we re-run the
same five-step pipeline---decompose, retrieve, verify, guide,
retry---with GLM-5.1 as the teacher and evaluate Qwen3-8B on the
already-retained $90$ Transfer-Set parents under the
independently written traces. Admission is not re-applied: every
parent stays in the set, so the scores below are a teacher-swap
ablation rather than a new filter.

Table~\ref{tab:alt-teacher} reports the result. GLM-5.1 guidance
unlocks $56$ of $90$ parents ($62.22$ $\mathrm{Par}@8$) and
$242$ of $280$ sub-problems ($86.43$ $\mathrm{Sub}@8$). The
DeepSeek V4 Pro traces remain $100$ by construction of the
filter. A majority of Transfer parents are therefore reachable
from the textbooks under two independently written guidance
traces, so the reachability witness is not an artifact of a
single teacher's style.

\begin{table}[h]
\centering
\small
\caption{Transfer-Set accuracy of Qwen3-8B under
textbook-grounded guidance written by two teachers. DeepSeek V4
Pro scores are $100$ by construction of the Naive Reachability
Filter. All numbers in \%.}
\label{tab:alt-teacher}
\begin{tabular}{lrr}
\toprule
Teacher & $\mathrm{Par}@8$ & $\mathrm{Sub}@8$ \\
\midrule
DeepSeek V4 Pro & 100.00 & 100.00 \\
GLM-5.1         &  62.22 &  86.43 \\
\bottomrule
\end{tabular}
\end{table}

Tables~\ref{tab:alt-teacher-source} and~\ref{tab:alt-teacher-disc}
break the parent-level result down by contest and by
sub-discipline. Coverage is broad rather than concentrated:
every source and every sub-discipline contributes at least one
solved parent.

\begin{table}[h]
\centering
\small
\caption{Qwen3-8B $\mathrm{Par}@8$ under GLM-5.1 guidance, by
olympiad source. All numbers in \%.}
\label{tab:alt-teacher-source}
\begin{tabular}{lrrr}
\toprule
Source & Parents & Passed & $\mathrm{Par}@8$ \\
\midrule
OPhO  &  1 &  1 & 100.00 \\
APhO  & 14 & 10 &  71.43 \\
NBPhO & 16 & 11 &  68.75 \\
IOAA  & 24 & 16 &  66.67 \\
IPhO  & 30 & 16 &  53.33 \\
EuPhO &  5 &  2 &  40.00 \\
\midrule
\textbf{Total} & \textbf{90} & \textbf{56} & \textbf{62.22} \\
\bottomrule
\end{tabular}
\end{table}

\begin{table}[h]
\centering
\small
\caption{Qwen3-8B $\mathrm{Par}@8$ under GLM-5.1 guidance, by
sub-discipline, using the same labels as
Table~\ref{tab:discipline}. All numbers in \%.}
\label{tab:alt-teacher-disc}
\begin{tabular}{lrrr}
\toprule
Sub-discipline & Parents & Passed & $\mathrm{Par}@8$ \\
\midrule
Relativity        &  2 &  2 & 100.00 \\
Thermal Physics   & 12 & 10 &  83.33 \\
Astrophysics      & 23 & 15 &  65.22 \\
Mechanics         & 28 & 17 &  60.71 \\
Electromagnetism  & 18 &  9 &  50.00 \\
Quantum Physics   &  4 &  2 &  50.00 \\
Waves             &  3 &  1 &  33.33 \\
\midrule
\textbf{Total} & \textbf{90} & \textbf{56} & \textbf{62.22} \\
\bottomrule
\end{tabular}
\end{table}

\section{An extracted record}
\label{app:extraction}

Each extracted record represents one physics problem in a unified
schema. The top-level fields capture provenance
(\texttt{source}, \texttt{source\_problem\_id}, \texttt{title},
\texttt{problem\_type}) and the full \texttt{problem} statement.
For multi-part problems, the record additionally holds a
\texttt{sub\_problems} list with one entry per sub-question; each
entry carries:
\begin{itemize}[topsep=0pt, partopsep=0pt, leftmargin=12pt, itemsep=-4.5pt]
  \item \texttt{problem\_id} --- a label such as \texttt{(a)} or
    \texttt{T1(b)} that disambiguates the sub-question within its
    parent;
  \item \texttt{problem} --- the sub-question text;
  \item \texttt{solution} --- the reference solution from the
    textbook or its official solution manual, where one is
    provided;
  \item \texttt{answer} --- the gold answer, with each final
    quantity wrapped in \verb|\boxed{}|;
  \item \texttt{answer\_type} --- one of the nine answer types
    listed in Section~\ref{sec:construction}, drawn from
    \{\texttt{NV}, \texttt{EX}, \texttt{EQ}, \texttt{TUP},
    \texttt{IN}, \texttt{MC}, \texttt{TF}, \texttt{QL}, \texttt{ALT}\};
  \item \texttt{type\_sequence} --- for composite types
    (\texttt{TUP} and \texttt{ALT}), the inner answer type at each
    position; empty for the seven primitive types.
\end{itemize}
Solo problems lift these per-sub-problem fields to the top level and
leave \texttt{sub\_problems} empty.

Below is one extracted record --- exercise P1.3
(``Force from a cone'') from Purcell's \emph{Electricity and
Magnetism} --- whose two sub-problems carry different answer types:
sub-problem (a) yields a single equation (\texttt{EQ}), and
sub-problem (b) yields an ordered tuple of two equations
(\texttt{TUP} with \texttt{type\_sequence = "EQ,EQ"}).

\begin{lstlisting}[
breakindent=0pt,
breakautoindent=false,
]
{
  "source": "Purcell_EM",
  "source_problem_id": "P1.3",
  "title": "Force from a cone",
  "problem_type": "Electromagnetic Fields",
  "problem": "(a) A charge $q$ is at the tip of     a hollow cone with surface charge density $     \\sigma$, slant height $L$, and half-angle $    \\theta$. ... (b) If the top half of the        cone is removed, what is the force on $q$,      and at what $\\theta$ is it maximum?",
  "sub_problems": [
    {
      "problem_id": "(a)",
      "problem": "A charge $q$ is at the tip of         a hollow cone with surface charge               density $\\sigma$, slant height $L$, and        half-angle $\\theta$. What can you say          about the force on $q$ due to the cone?",
      "solution": "Integrating the per-ring             vertical force over $x \\in (0, L]$             gives $F \\propto \\int_{0}^{L} dx/x$,          which diverges as $x \\to 0$.",
      "answer": "\\boxed{F = \\infty}",
      "answer_type": "EQ",
      "type_sequence": ""
    },
    {
      "problem_id": "(b)",
      "problem": "If the top half of the cone is        removed, what is the force, and at what         $\\theta$ is it maximum?",
      "solution": "The integral now runs from $L        /2$ to $L$: $F = \\frac{q\\sigma\\ln 2}         {4\\epsilon_0}\\sin 2\\theta$, which is         maximised at $2\\theta = 90^\\circ$.",
      "answer": "\\boxed{F = \\frac{q\\sigma\\ln        2} {4\\epsilon_0} \\sin 2\\theta},              \\boxed{\\theta_{\\max} = 45^\\circ}",
      "answer_type": "TUP",
      "type_sequence": "EQ,EQ"
    }
  ]
}
\end{lstlisting}

\section{LLM-judge prompt}
\label{app:judge}

For the verifier's LLM-judge fallback (Section~\ref{sec:metrics}), we
ask DeepSeek-V4-Flash-0731 to compare a student-extracted answer against
the reference answer and return a binary verdict together with a
short justification. The full prompt template is reproduced below.
The placeholders \verb|{{problem}}|, \verb|{{RS}}|, \verb|{{RA}}|,
\verb|{{SS}}|, \verb|{{SA}}| are filled in at run-time with the
problem statement, the reference solution, the reference final
answer(s), the student's full solution, and the answer(s) extracted
from the student's solution, respectively.

\begin{lstlisting}[
breakindent=0pt,
breakautoindent=false,
]
You are an expert physics / astronomy grader. You will be given

1. The original problem statement.
2. The reference (official) step-by-step solution.
3. The reference final answer(s).
4. A student's full solution.
5. The answer(s) extracted from the student's solution.

Your job is ONLY to decide whether the student's final answer is mathematically and physically equivalent to the reference final answer, taking into account acceptable tolerances (e.g. unit conversion, reasonable rounding, equivalent algebraic forms, equivalent sign conventions, equivalent multiple-choice letters, etc.).

You are not grading the derivation; if the final answer matches up to acceptable numerical tolerance or algebraic equivalence, it is correct, even if the reasoning was imperfect. Conversely, a correct derivation that ends with an incorrect numerical value should be judged incorrect.

You MUST output your response in exactly the following format, with nothing else before or after:

## Equivalence Judgement
TRUE

## Justification
<one or two sentences explaining the decision>

Where the line after "## Equivalence Judgement" is either the single word TRUE or the single word FALSE (uppercase, no punctuation).

Guidelines for numerical answers:
- Accept answers within ~2% relative tolerance for "order of magnitude" / "estimate" style problems.
- Accept answers within the explicitly stated tolerance range given in the reference solution, if any.
- Different but equivalent units (e.g. "20 kT" vs "2e4 T", "90 m" vs "90 meters") are equivalent.
- Different precisions that round to the same value at the reference's stated precision are equivalent.

Guidelines for symbolic answers:
- Expressions that are algebraically equal after simplification are equivalent.
- Expressions differing only by a named physical constant that has been replaced with its symbol (e.g. "c" vs "3e8 m/s") are equivalent.
- Expressions differing only by dimensionally-trivial rearrangement (e.g. d * sqrt(d/GM) vs sqrt(d^3/GM)) are equivalent.

Guidelines for tuples / multi-answer:
- Match element-wise; order matters unless the problem clearly says otherwise.
- For ALT answer types (alternative acceptable answers), matching ANY one of the listed acceptable answers is sufficient.

Guidelines for True/False / multiple choice:
- Normalize case and synonyms ("Yes" == "True" == "T", "No" == "False" == "F").

Guidelines for qualitative answers (short natural-language phrases such as "tidal forces", "general relativity", "spectral line broadening"):
- Accept answers that name the same physical concept / mechanism / object as the reference, even if the wording differs (e.g. "tidal force" vs "tidal forces" vs "differential gravitational pull from the companion").
- Accept reasonable synonyms and equivalent technical terms used in the field.
- Reject answers that name a different concept, even if related (e.g. "radiation pressure" is NOT equivalent to "tidal forces").
- A correct answer buried inside a longer explanation is still correct as long as the named concept is unambiguously the reference one.
- Spelling, capitalization, pluralization and trivial word-order differences are immaterial.

Do not include any text outside the two required sections.

--- Problem ---
{{problem}}

--- Reference solution ---
{{RS}}

--- Reference final answer(s) ---
{{RA}}

--- Student's full solution ---
{{SS}}

--- Answer(s) extracted from student's solution ---
{{SA}}
\end{lstlisting}

\section{Verifier consistency}
\label{app:verifier}

Section~\ref{sec:metrics} scores each sub-problem with a two-stage
verifier: the type-aware rule-based judger first, then
DeepSeek-V4-Flash-0731 on every residual the rules reject. We adopt the
conservative assumption that every rule-based accept would also be
accepted by the judge model, and measure consistency on the
$6{,}720$ attempt-level sub-problem judgements from Qwen3-8B's
$24$-sample Transfer-Set run ($90$ parents, $280$ unique
sub-problems).

Table~\ref{tab:verifier-agree} reports the resulting
$2{\times}2$. The two stages agree on $4{,}967$ of $6{,}720$
judgements ($73.91\%$). The $739$ rule-based accepts
($11.00\%$) are treated as joint accepts. Of the $5{,}981$
residuals routed to the LLM, $4{,}228$ ($70.69\%$) remain
incorrect and $1{,}753$ ($29.31\%$) are recovered as equivalent.
Treating the cascade---and the assumption above---as the reference
label, the rule-based judger therefore has precision $1$ and recall
$29.65\%$: it is a high-precision, low-recall filter. This is why
we expose only the rule-based stage as an RL reward and reserve
the LLM fallback for leaderboard evaluation.

\begin{table}[h]
\centering
\small
\caption{Attempt-level agreement between the rule-based judger and
the LLM fallback on Qwen3-8B Transfer-Set rollouts. A rule-based
accept is assumed to be an LLM accept (starred cell, unobserved).}
\label{tab:verifier-agree}
\begin{tabular}{lrr}
\toprule
 & LLM accept & LLM reject \\
\midrule
Rule accept & 739 & $0^{\ast}$ \\
Rule reject & 1{,}753 & 4{,}228 \\
\midrule
Total & 2{,}492 & 4{,}228 \\
\bottomrule
\end{tabular}
\end{table}

Disagreement is one-sided and type-dependent. Discrete types like MC and TF agree almost
perfectly. Recovered
cases concentrate on \texttt{EQ}, \texttt{IN}, \texttt{NV},
\texttt{TUP}, and \texttt{QL}, where the LLM accepts algebraically
equivalent rearrangements, unit or constant synonyms ($k$ vs.\
$k_B$), interval-versus-inequality notation, numerical values
within tolerance, and short qualitative aliases that the symbolic
matcher rejects.

\section{Application Set redaction from the training material}
\label{app:redaction}

Because Application Set problems are themselves end-of-chapter
exercises drawn from textbooks in $\mathcal{C}$, naively releasing
those textbooks as training material would leak every Application
Set answer. We therefore physically redact, from the raw markdown
that becomes $\mathcal{C}$, both the problem statement and the
reference solution of every retained Application Set parent. The
process runs in two passes followed by an audit.

\vspace{+2pt}\noindent\textbf{Pass 1: problem statements.}
For each retained Application Set parent, we locate its problem
block in its source textbook and replace the block with a
single-line marker of the form
\nolinkurl{<!-- [TEST-SET REDACTION] source_problem_id=<id> source=<src> -->}. 

Block boundaries are determined by source-specific layout rules, since the eight contributing textbooks use heterogeneous exercise
layouts.

\vspace{+2pt}\noindent\textbf{Pass 2: solutions and answer keys.}
A second pass excises the corresponding solutions and answer keys,
which for several textbooks live in entirely separate files:
Purcell's \nolinkurl{electricity\_and\_magnetism/12\_solutions/full.md}
aggregates every chapter-end solution; Kleppner ships a parallel
\nolinkurl{introduction\_to\_mechanics\_solution\_manual\_by\_chapter}
directory; each chapter of Morin ends with its own
``\nolinkurl{X.5 Solutions}'' section; French's
\nolinkurl{11\_answers/full.md} lists short answers chapter-by-chapter;
Taylor \& Wheeler's \nolinkurl{10a\_answers/full.md} concatenates every
chapter's answers into a single paragraph, from which we surgically
remove only the spans corresponding to redacted parents; Eisberg \&
Resnick's inline back-of-chapter answer paragraphs are trimmed
similarly. For the remaining contributors (Carroll \& Ostlie and
Palen's \emph{Schaum's Astronomy}), the textbook provides no
separate solution section, so Pass 1 alone suffices.

\vspace{+2pt}\noindent\textbf{Audit.}
After both passes, we walk every \nolinkurl{full.md} in the corpus,
extract 40-character normalised anchors from each retained parent's
problem and solution texts, and search for verbatim matches. The
assembled corpus $\mathcal{C}$ contains zero such matches; the same
audit additionally confirms that every retained Application Set
parent has at least one redaction marker placed in the corpus. The
full redaction and audit scripts are released with the instruction
layers; the Corpus itself is not redistributed
(Appendix~\ref{app:release}).

\section{Convergence curves for the remaining methods}
\label{app:curves}

\begin{figure}[h]
  \centering
  \includegraphics[width=\columnwidth]{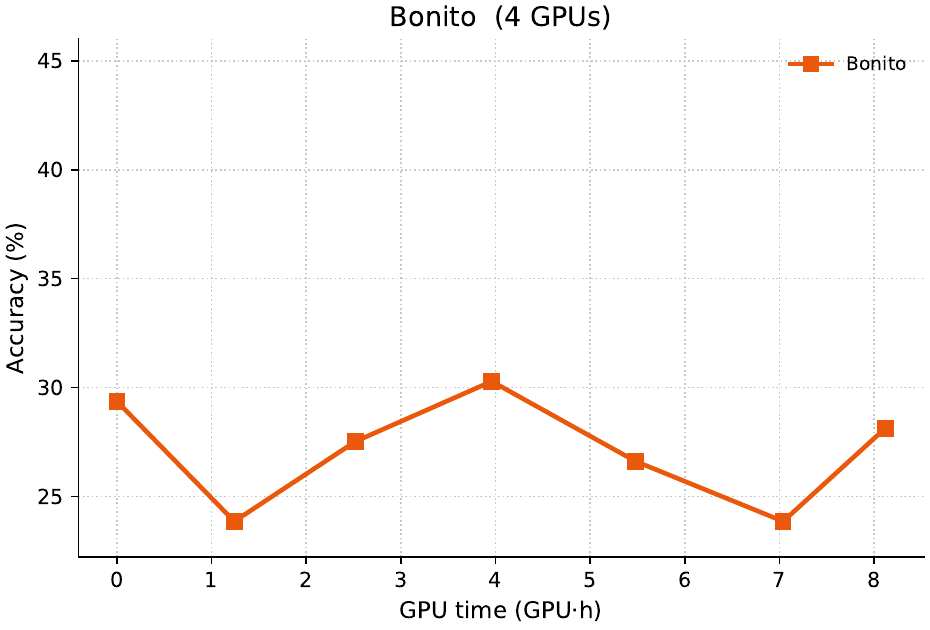}
  \caption{Application-Set $\mathrm{Sub}@8$ of \textbf{Bonito}
  against cumulative GPU-time (4$\times$A800). The curve
  oscillates between $\approx 24\%$ and $\approx 30\%$ over
  $8.12$ GPU hour and ends at $28.13$, below the Qwen3-8B
  baseline, with no sustained climb.}
  \label{fig:bonito-curve}
\end{figure}

\begin{figure}[h]
  \centering
  \includegraphics[width=\columnwidth]{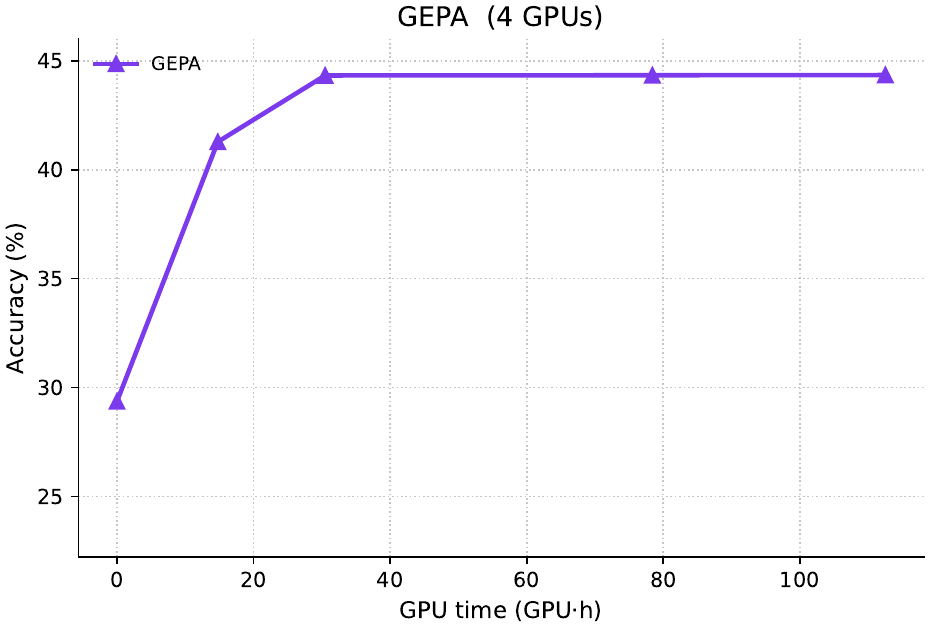}
  \caption{Application-Set $\mathrm{Sub}@8$ of \textbf{GEPA}
  against cumulative GPU-time (4$\times$A800). GEPA reaches
  $44.3\%$ by $30.5$ GPU hour and remains there through the
  $112.5$ GPU hour endpoint, matching the $44.34$ in
  Table~\ref{tab:main}.}
  \label{fig:gepa-curve}
\end{figure}

\begin{figure}[h]
  \centering
  \includegraphics[width=\columnwidth]{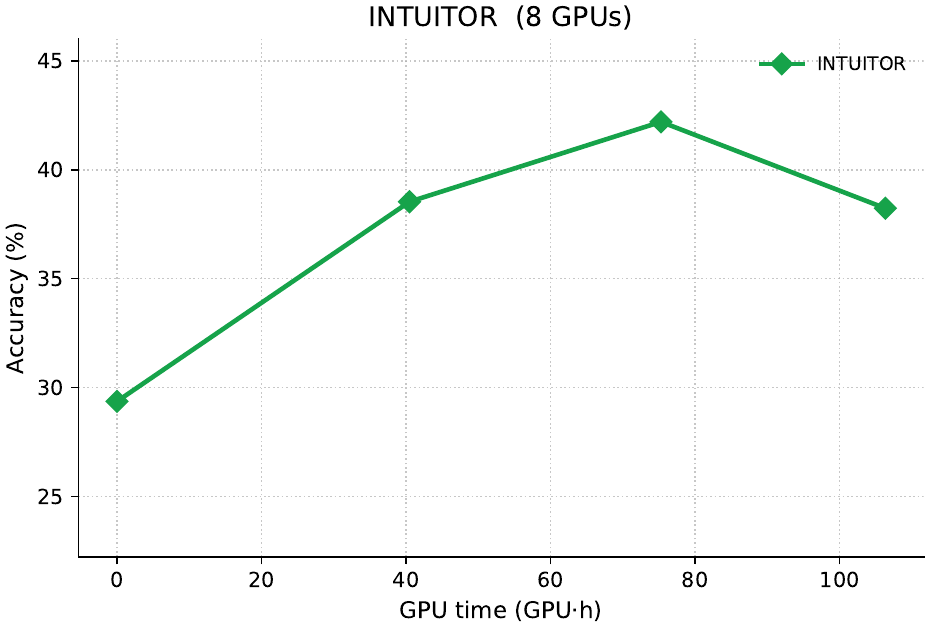}
  \caption{Application-Set $\mathrm{Sub}@8$ of
  \textbf{Intuitor} against cumulative GPU-time (8$\times$A800).
  Intuitor rises to $\approx 42\%$ at $75.3$ GPU hour and then
  drops to $38.23$ at the $106.3$ GPU hour endpoint.}
  \label{fig:intuitor-curve}
\end{figure}

\begin{figure}[h]
  \centering
  \includegraphics[width=\columnwidth]{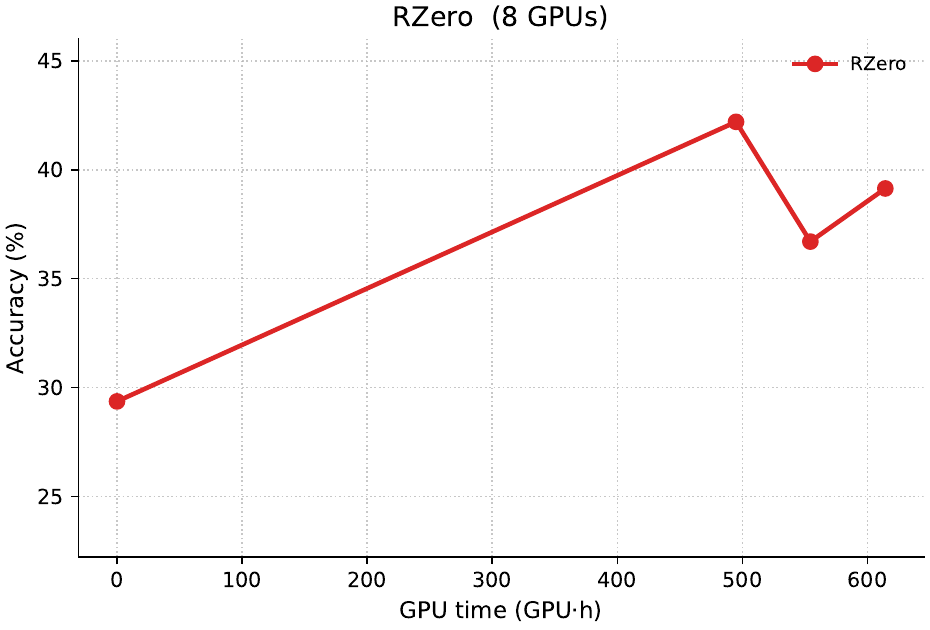}
  \caption{Application-Set $\mathrm{Sub}@8$ of
  \textbf{R-Zero} against cumulative GPU-time (8$\times$A800).
  R-Zero spends $495$ GPU hour to peak near $42\%$, then
  drops and ends at $39.14$ after $614$ GPU hour.}
  \label{fig:rzero-curve}
\end{figure}

Figure~\ref{fig:ace-curve} in the main text plots ACE's
Application-Set $\mathrm{Sub}@8$ against cumulative GPU-time,
averaged over three seeds. The corresponding curves for the
other four profiled methods are reproduced below
(Figures~\ref{fig:bonito-curve}--\ref{fig:rzero-curve}). Each
panel uses a method-specific GPU-time axis; the accuracy axis
is fixed at $25$--$45\%$ for visual comparison. GEPA climbs
then sits flat; Intuitor and R-Zero peak and decline; Bonito
never leaves the baseline band.

\section{Replication detail}
\label{app:budgets}

We reproduce each baseline from its official repository and adapt it
to StudyBench's training-material layers and evaluation protocol.
Unless noted otherwise, all changes are implemented by subclassing
upstream base classes or adding thin wrapper scripts. Below we summarise the adaptations for each
method.

\vspace{+2pt}\noindent\textbf{GEPA.}
GEPA consumes the \emph{Instructions with Answer} layer and evolves a
system prompt via reflective genetic search. We made four adaptations:
\begin{itemize}[topsep=0pt, partopsep=0pt, leftmargin=12pt, itemsep=-4.5pt]
  \item \textbf{Train/validation split.} We wrote a data loader that
    shuffles the instruction set and reserves $32$ items as a
    validation set, using the remainder for training. The validation
    set is kept small because GEPA re-evaluates it after every
    iteration; a larger set slows each rollout substantially and, in
    our runs, occasionally caused the evaluation loop to hang for
    unknown reasons.
  \item \textbf{StudyBench adapter.} We subclass GEPA's \texttt{Adapter}
    class, which is responsible for generation, scoring, and returning
    textual feedback. Our adapter mirrors the benchmark's multi-turn
    generation protocol and its rule-plus-LLM verifier, and calls
    Qwen3-8B to produce the reflective feedback GEPA uses to mutate
    candidate system prompts.
  \item \textbf{Checkpoint callback.} We register a callback that
    snapshots the best system prompt seen so far every $100$ rollouts.
    This is a bookkeeping convenience for post-hoc inspection and does
    not change the search objective or final result.
  \item \textbf{Non-invasive integration.} All of the above are
    implemented through inheritance; the upstream GEPA package itself
    is left unmodified.
\end{itemize}

\vspace{+2pt}\noindent\textbf{Bonito.}
Bonito consumes the \emph{Corpus} layer to synthesise instruction
data and fine-tunes the base model on it. We made three adaptations:
\begin{itemize}[topsep=0pt, partopsep=0pt, leftmargin=12pt, itemsep=-4.5pt]
  \item \textbf{Corpus chunking.} We wrote a data loader that splits
    each textbook in $\mathcal{C}$ into $2{,}048$-token chunks for
    downstream instruction extraction.
  \item \textbf{Physics-problem task type.} We added a
    \emph{physics-problem} extraction category to Bonito's
    task-conditioned generator. A dedicated prompt asks the model to
    mine standalone physics exercises from textbook passages.
  \item \textbf{Self-evolution loop.} The original Bonito pipeline
    stops after instruction extraction. We add one further
    round: instructions produced by Qwen3-8B are fed back to train the
    same model, closing a single self-evolution cycle on the corpus.
\end{itemize}

\vspace{+2pt}\noindent\textbf{ACE.}
We make three classes of changes to the open-sourced ACE codebase.
\emph{(i) Backbone.} All three roles (Generator, Reflector, Curator) are driven by a single locally-served Qwen3-8B; we strip \texttt{<think>} traces between roles, disable thinking on Reflector/Curator (kept on the Generator), and harden the JSON parser against LaTeX escape sequences.
\emph{(ii) Playbook management.} We cap the playbook at $12{,}000$ tokens and implement the token-budget Pruning Trigger described in the ACE paper but absent from the released code, which drops the lowest-utility bullets whenever the budget is exceeded; we additionally enable the bulletpoint analyzer for embedding-based deduplication and LLM-driven merging of near-duplicate entries.
\emph{(iii) Evaluation.} Rather than use ACE's built-in \texttt{eval\_only} (whose prompt format, judge, and metric are not aligned with StudyBench), we leave StudyBench's evaluator untouched and inject the playbook into its Generator system message, reducing the ACE-vs-baseline comparison to a single controlled variable.

\vspace{+2pt}\noindent\textbf{EvoSkill.}
EvoSkill consumes the \emph{Instructions with Answer} layer and evolves a
Claude Code agent program---a system prompt together with a folder of
reusable skills---via a failure-driven proposer/generator/evaluator loop.
Unlike GEPA, which revises a single instruction in place, each iteration
can add or edit skill files and is accepted only if it improves held-out
accuracy. We preserve the official loop (skill-only mutations, a
size-$3$ frontier) and make five adaptations:
\begin{itemize}[topsep=0pt, partopsep=0pt, leftmargin=12pt, itemsep=-4.5pt]
  \item \textbf{Instruction-pool loader.} We convert the StudyBench
    verifiable instruction JSONs into EvoSkill's CSV layout. Only items
    with a gold answer are kept; multi-turn parents are flattened into
    conversational prompts that match the benchmark protocol. Because
    the pool mixes two source files (single-ask and multi-sub), we
    partition \emph{per source} at $20$/$10$/$70$
    train/val/held-out ($283$/$141$ train/val rows) rather than using
    EvoSkill's built-in stratified split; the loader honours the
    pre-computed \texttt{split} column so the remaining labelled
    instructions are never seen during evolution, matching the original
    paper's small-train convention.
  \item \textbf{Skill-category clustering.} The official pipeline first
    clusters the dataset into $K$ skill categories with an LLM
    classifier and then round-robins failure samples by category. We
    keep that design but replace the OfficeQA taxonomy with eight
    physics-reasoning labels (conservation laws, force/dynamics,
    differential equations, fields/potentials, thermodynamics,
    waves/optics, quantum/modern, and estimation/dimensional analysis),
    assigned by a DSPy classifier. Sampling therefore rotates over
    reasoning skills rather than answer types or textbook topics.
  \item \textbf{Frontier parent selection.} The released runtime
    defaults to greedy \texttt{best} selection, which never mutates the
    $2$nd/$3$rd-best lineages when the frontier has size greater than
    one. We expose the paper's \texttt{round\_robin} strategy in the
    project config so every frontier member gets equal mutation budget.
  \item \textbf{Rule-based physics verifier.} We replace EvoSkill's
    default \texttt{multi\_tolerance} string/numeric scorer with the
    same type-aware, rule-only judger used by the benchmark
    (Section~\ref{sec:metrics}), with no LLM-judge fallback. Composite
    types (\texttt{TUP}/\texttt{ALT}) are graded slot-by-slot through
    the declared \texttt{type\_sequence}. Because the scorer signature
    carries no type field, we rebuild a question-to-schema lookup from
    the CSV. Each judgment runs in a \texttt{spawn} subprocess with a
    hard timeout so a pathological \texttt{sympy} input cannot stall
    the event loop---the same isolation used for GRPO, TTRL, and
    R-Zero.
  \item \textbf{Prompt and evaluation.} We rewrite the default OfficeQA
    system prompt as a physics-olympiad solver that must emit one
    \verb|\boxed{}| per final answer in valid \LaTeX{}, matching the
    grader's extraction rule. Training uses Claude Code with a tool
    loop; StudyBench's evaluator is a raw Messages API with no tools.
    Rather than use EvoSkill's built-in \texttt{eval} (whose prompt
    format, judge, and metric are not aligned with StudyBench), we
    export the best frontier program by inlining the task description
    and the learned skills---filtering out Claude Code's stock
    meta-skills---into a single system-prompt blob, and inject that
    blob as the entire system message. This reduces the
    EvoSkill-vs-baseline comparison to a single controlled variable,
    the same evaluation pattern used for GEPA and ACE. We run only on
    Opus~4.7 with Claude Code, budget $\sim 1.5$ epochs over the
    $283$-row train pool (the convention of the original paper), and
    stop after $5$ iterations without improvement. All of the above
    are thin wrappers (a preprocess script, a scorer hook, a
    split-honouring loader, and an export script); the upstream loop
    itself is left unmodified.
\end{itemize}

\vspace{+2pt}\noindent\textbf{GRPO.}
GRPO consumes the \emph{Instructions with Answer} layer and trains
with group-relative policy optimisation under a gold outcome
reward. We keep verl's official loop---group-normalised advantages
over $n$ rollouts, a low-variance KL penalty, and no learned
critic---and make four adaptations:
\begin{itemize}[topsep=0pt, partopsep=0pt, leftmargin=12pt, itemsep=-4.5pt]
  \item \textbf{Instruction-pool loader.} We convert the StudyBench
    instruction JSONs into verl parquet. Gold answers are stored as
    \texttt{reward\_model.ground\_truth} and enter the $0/1$ outcome
    reward on every actor update. Multi-turn parents are flattened
    into conversational histories that match the benchmark protocol,
    and a short suffix on the last user turn asks for one
    \verb|\boxed{}| per final answer. We shuffle the $1{,}420$
    labelled items and hold out $5\%$ for validation.
  \item \textbf{Rule-based physics verifier.} We replace verl's
    GSM8K/MATH grader with the same type-aware, rule-only judger used
    by the benchmark (Section~\ref{sec:metrics}), with no LLM-judge
    fallback. Composite types (\texttt{TUP}/\texttt{ALT}) are graded
    slot-by-slot through the declared \texttt{type\_sequence}. A
    worker subprocess stages an ANTLR~4.11 runtime for
    \texttt{sympy.parsing.latex} so the parent can keep the 4.9
    runtime that Hydra/OmegaConf require---the same isolation used
    for TTRL and R-Zero. Each judgment is hard-timeout-bounded so a
    pathological \texttt{sympy} input cannot stall a training step.
  \item \textbf{Thinking-mode generation.} Qwen3 thinking is required
    for the physics RL signal to be useful: we enable it and raise
    the response cap from verl's default $2{,}048$ to $16{,}384$
    tokens. A first run at $8{,}192$ truncated $57\%$ of
    completions, collapsing the reward.
  \item \textbf{Scale and evaluation.} The labelled pool is small
    ($\sim 1.3$k train rows), so we use a train batch of $128$
    ($\sim 10$ groups per epoch) and a group size of $n{=}8$ so
    that sparse $0/1$ physics rewards still yield a usable
    within-group baseline. We train Qwen3-8B and
    Llama-3.2-3B-Instruct for $15$ epochs on a single
    $8{\times}$A800-80GB node. Mid-training validation uses the
    held-out split under the rule-based verifier only; final
    evaluation uses StudyBench's standard protocol, not verl's
    built-in GSM8K/MATH eval. All of the above are thin wrappers
    (a preprocess script, a \texttt{custom\_reward\_function} hook,
    and a launch script); the upstream verl package itself is left
    unmodified.
\end{itemize}

\vspace{+2pt}\noindent\textbf{Intuitor.}
Intuitor consumes the \emph{Instructions without Answer} layer and
trains with GRPO under the policy's own self-certainty as reward.
We made two adaptations:
\begin{itemize}[topsep=0pt, partopsep=0pt, leftmargin=12pt, itemsep=-4.5pt]
  \item \textbf{StudyBench training split.} We load our own instruction
    set and partition it into training and validation splits for the
    GRPO run. Because Intuitor is designed to train without an
    external reward signal, no verifier is used during optimisation.
  \item \textbf{Validation-time monitoring.} On the held-out validation
    split we attach the same rule-based verifier used in benchmark
    evaluation (Section~\ref{sec:metrics}) to track learning curves and
    guard against degenerate training trajectories. To keep validation
    lightweight, we expose only the rule-based component and omit the
    LLM-judge fallback.
\end{itemize}

\vspace{+2pt}\noindent\textbf{TTRL.}
TTRL consumes the \emph{Instructions without Answer} layer and trains
with GRPO under majority-vote consistency as reward. We keep the
official loop---vote a pseudo-label over $n_{\mathrm{votes}}$
rollouts, then update the actor on $n_{\mathrm{samples}}$ of
them---and make four adaptations:
\begin{itemize}[topsep=0pt, partopsep=0pt, leftmargin=12pt, itemsep=-4.5pt]
  \item \textbf{Instruction-pool loader.} We convert the StudyBench
    instruction JSONs into TTRL parquet. Gold answers, when present,
    are stored only for the offline \texttt{label\_accuracy} metric;
    \texttt{apply\_ttrl\_gt} overwrites \texttt{ground\_truth} with
    the majority-voted pseudo-label before every actor update, so
    labels never enter the loss. Multi-turn parents are flattened
    into conversational histories that match the benchmark protocol,
    and a short suffix on the last user turn asks for one
    \verb|\boxed{}| per final answer.
  \item \textbf{Physics-aware majority vote.} Upstream TTRL extracts
    the last \verb|\boxed{}| and pre-simplifies it with
    \texttt{sympy}, which is appropriate for single-number math
    answers. Physics composites (\texttt{TUP}/\texttt{ALT}) carry
    several boxes; voting on only the last one collapses the answer
    so the pseudo-label is ungradable. We therefore vote on the full
    ordered tuple of normalised boxes for composite types and on the
    last box for atomic or unknown types, drop the inline
    \texttt{sympy.simplify} (fragile under units and scientific
    notation), and reconstruct a multi-box pseudo-GT that
    round-trips through the same extractor the reward uses.
  \item \textbf{Rule-based physics verifier.} We replace
    \texttt{ttrl\_math}'s mathd/sympy grader with the same
    type-aware, rule-only judger used by the benchmark
    (Section~\ref{sec:metrics}), with no LLM-judge fallback. A
    worker subprocess stages an ANTLR~4.11 runtime for
    \texttt{sympy.parsing.latex} so the parent can keep the 4.9
    runtime that Hydra/OmegaConf require---the same isolation used
    for R-Zero. Each judgment is hard-timeout-bounded.
  \item \textbf{Scale and evaluation.} Because the instruction pool
    is $\sim 70\times$ larger than AIME, we reduce the official
    $64$/$32$ vote/train split to $16$/$8$ so that rollouts per
    step stay comparable, and we raise the response cap to
    $16{,}384$ tokens to leave room for Qwen3 thinking traces. We
    train Qwen3-8B and Llama-3.2-3B-Instruct for $3$ epochs on a
    single $8\times$A800-80GB node. Mid-training validation uses a
    held-out textbook split under the rule-based verifier only;
    final evaluation uses StudyBench's standard protocol, not
    TTRL's built-in AIME-style eval.
\end{itemize}

\vspace{+2pt}\noindent\textbf{R-Zero.}
We adapt R-Zero~\citep{huang2025rzero} to our setting
without altering its core algorithm: we preserve the Challenger/Solver GRPO
objectives, the uncertainty reward and BLEU diversity penalty,
majority-vote pseudo-labels with $m=10$ Solver samples per question, the
$\hat{p}\in[0.3,0.8]$ consistency band, and the co-evolution loop, and
only rewrite (i)~the Challenger and Solver system prompts from math
problem setting/solving to physics-olympiad-style problem setting/solving,
enforcing a single boxed numerical, symbolic, or equation answer with no
units, no proofs, and no plots; (ii)~the verifier, replacing
\texttt{mathruler.grade\_answer} with the same rule-based, type-aware
judger used by our benchmark (no LLM fallback), with a monkey-patch that
routes \texttt{sympy.parsing.latex.parse\_latex} through
\texttt{latex2sympy2\_extended} to sidestep the \texttt{omegaconf}/\texttt{sympy}
ANTLR version conflict; and (iii)~some pipeline plumbing (local Parquet
instead of a HuggingFace round-trip, dynamic checkpoint-path resolution
that tolerates mid-training crashes, and \texttt{set -e} propagation
across stages). We use \texttt{Qwen3-8B} as the base model
and run 3 co-evolution iterations (the saturating regime reported in the
original paper), with 6 Challenger and 20 Solver GRPO steps per iteration
and \texttt{max\_response\_length}${=}8192$, on a single 8$\times$A800-80GB
node; one full run takes approximately ten days of wall time. Final
evaluation uses our benchmark's standard protocol, not R-Zero's built-in
evaluation.

\section{LLM Usage}
\label{secllm_usage}

During preparation of this manuscript we used large language models (LLMs) as editorial and data-processing tools to support the writing and dataset preparation workflow. Their assistance included PDF-to-Markdown conversion and automated correction of common OCR and formatting errors. Also we use it with editorial help on the manuscript text such as grammar and style polishing, sentence rephrasing, paragraph restructuring, and drafting initial versions of descriptive passages. Crucially, all LLM-assisted outputs were treated as drafts, every suggestion was reviewed, edited, and approved by the authors before incorporation. We disclose this use here and in the submission form in accordance with community guidance.

\end{document}